\documentclass[conference]{IEEEtran}
\IEEEoverridecommandlockouts
\usepackage{cite}
\usepackage{amsmath,amssymb,amsfonts}
\usepackage{algorithm}
\usepackage[hidelinks]{hyperref}
\usepackage{algpseudocode}
\usepackage{graphicx}
\usepackage{textcomp}
\usepackage{xcolor}

\newcommand{\method}{MIRACLE }
\usepackage{graphicx}
\graphicspath{{./figures/}}

\def\BibTeX{{\rm B\kern-.05em{\sc i\kern-.025em b}\kern-.08em
    T\kern-.1667em\lower.7ex\hbox{E}\kern-.125emX}}
\begin{document}

\title{\method: Inverse Reinforcement and Curriculum Learning Model for Human-inspired Mobile Robot Navigation}

\author{Nihal Gunukula, Kshitij Tiwari and Aniket Bera%
\thanks{The authors are with the Department of Computer Science, Purdue University, USA,
        {\tt\small \{ngunukul,tiwarik,ab\}@purdue.edu}}%
}

\maketitle

\begin{abstract}
In emergency scenarios, mobile robots must navigate like humans, interpreting stimuli to locate potential victims rapidly without interfering with first responders. Existing socially-aware navigation algorithms face computational and adaptability challenges. To overcome these, we propose a solution, \method- an inverse reinforcement and curriculum learning model, that employs gamified learning to gather stimuli-driven human navigational data. This data is then used to train a Deep Inverse Maximum Entropy Reinforcement Learning model, reducing reliance on demonstrator abilities. Testing reveals a low loss of 2.7717 within a 400-sized environment, signifying human-like response replication. Current databases lack comprehensive stimuli-driven data, necessitating our approach. By doing so, we enable robots to navigate emergency situations with human-like perception, enhancing their life-saving capabilities.

\end{abstract}

\section{Introduction}
Mobile robots are being increasingly deployed in the workplace  \cite{kumar_robot_2004, helge_seneka_2012, schilling_remote_2005}. In these emergencies, the robots need to deal with partially-observable scenes- the location of the victim is unknown and it is up to the first responder to interpret the signs of life and localize the victim as accurately and rapidly as possible. Having a mobile robot perform this task would mean that the mobile robot must be able to respond to stimuli in a similar fashion as human beings. To do this, mobile robots need to employ adaptive and socially-aware navigation, i.e., the ability to interpret stimuli and navigate to the goal accordingly.

The traditional approaches to address adaptive socially-aware navigation were to utilize mathematical models \cite{araceli_socially_2017}, social forces \cite{gonzalo_robot_2013}, and probabilistic models\cite{meera_socially_17}. These methods work by simulating human social navigation, which is then transferred to a mobile robot. With the recent advances in data-driven methods, machine learning has been the preferred approach because it allows the algorithm to learn, be flexible in new environments, and make real-time decisions while being generalizable across myriad conditions. In order to utilize machine learning, human navigation data, i.e., a compilation of trajectories that an individual takes in a given environment, is necessary to train the algorithm. The issue with this is that this data is typically gathered through observing real-world interactions and decisions which is tedious when trying to replicate a specific scenario to research. 

\begin{figure}[!htbp]
    \centering
    \includegraphics[width=\columnwidth]{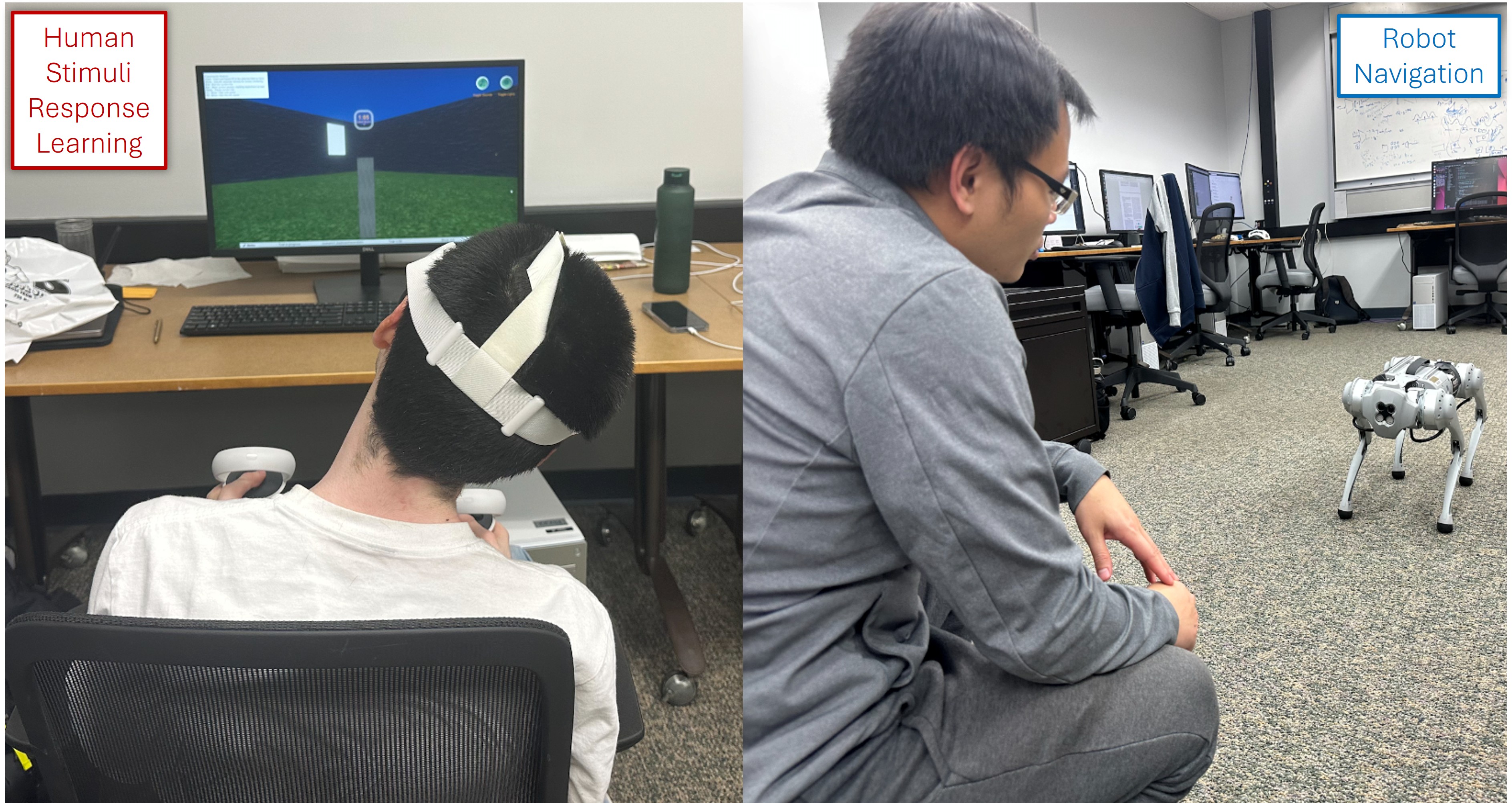}
    \caption{\textit{Illustrating the virtual environment (left) built for \textbf{MIRACLE}, alongside the UniTree GO1 mobile robot (right) that will replicate the stimuli-based navigation behavior in the real world.}}
    \label{fig:motivation}
\end{figure}

This method of real-world data collection also suffers from the fact that factors like urgency, prior understanding of the environment, comfort around other people, etc., can influence navigational decisions, which can make it challenging to decipher which factors led to the resultant behavior.

Another challenge with developing stimuli-driven navigation is the impact that demonstrators' biases can have because a demonstrator's poor decision, like colliding with a stationary object, would often be passed down to the mobile robot \cite{hurtado_learning_2021}. These biases can make the human navigational data convoluted for the algorithm, making it unable to discern what it should take away from the demonstration.

To address this demonstrator's bias issue, numerous learning techniques have been used, like the learning-relearning framework \cite{hurtado_learning_2021} and the adversarial inverse reinforcement model \cite{fu_learning_2018}. However, they both are too computationally expensive in comparison to the maximum entropy inverse reinforcement model, which uses entropy to embed variability in the algorithm to reduce demonstrator-induced bias\cite{asselmeier_evolutionary_2023}.

On top of the aforementioned challenges, there is also a lack of real-world testing of stimuli-driven navigation algorithms on mobile robots \cite{fahad_learning_2018, kingma_comprehensive_19, asselmeier_evolutionary_2023}. Real-world testing is important because the real world is relatively more complex than simulation \cite{grim_how_2013}. 

In this paper, we make the following novel contributions:
\begin{itemize}
\item To \textbf{easily gather large amounts of human stimuli-driven navigational data}, and allowing researchers to easily gather data for diverse scenarios, we utilize gamified learning to gather data regarding people's response to stimuli in a controlled setting as shown in Fig.~\ref{fig:motivation} (left). 
\item To \textbf{limit the impact of demonstrators' bias}, we developed a deep maximum entropy inverse reinforcement model to benefit from the variance that maximum entropy embeds into the algorithm while using deep learning to ensure that the algorithm does not stray too far from its intended goal. 
\item {When training the machine learning model, we provide it with the best-collected data we have first and then allow the data to get progressively worse. This is done as the data is utilized first from the later trials, after which the demonstrator has reached their peak of accomplishing the goal-finding task, and then gives it the data of the earlier trials.}
\end{itemize}

\section{Related Works}
In this section, we discuss prior works on adaptive socially-aware navigation for mobile robots.

\subsection{Stimuli-driven navigation for mobile robots}
When developing a stimuli-driven navigation algorithm for mobile robots, one can choose between a \textit{probabilistic},\textit{theory based}, and \textit{data-driven} approach.

One such probabilistic approach is the nonlinear optimization model which serves to illustrate behavior that would allow a mobile robot to be integrated into a workplace \cite{banisetty_optimization_2021}. This method has shown promise in a similar navigation technique, socially aware navigation. The PaCeT system presented utilizes linear optimization to navigate through hallways. The probabilistic approach identifies itself, by utilizing the probability of actions and outcomes for the mobile robot to make decisions \cite{sebastin_probabilistic_2000, cummins_probabilistic_07, salma_probabilistic_22}.

Another approach has been theory-based approaches which utilize a specific socially aware navigation theory for the mobile \cite{martinez_theory_2014, kadar_field_1998, moyle_theory_03}. These approaches work by using theoretical mathematical models to represent the navigation issue of the mobile robot. This mathematical model is then solved and the solution is provided to the mobile robot to implement.

Both the probabilistic approach and the theory-based approach struggle with being scenario-specific, too computationally intensive, extremely hard to scale, and challenging to balance the trade-off between safety and efficiency. 

Machine learning, a data-driven approach, has been promising at implementing stimuli-driven navigation in reinforcement machine learning, specifically deep reinforcement learning, and inverse reinforcement learning because of their learning abilities given policy reward functions replicating human navigation behavior \cite{akalin_reinforcement_2021}. Deep Reinforcement Learning has been utilized in robot navigation research for its unique architecture \cite{hoage_survey_20}. However, it struggles as it does not leverage human navigation making it reliant on the policy function given to it by the researcher \cite{asselmeier_evolutionary_2023}. To account for this limitation, Inverse Reinforcement Learning can be used because of its capability to develop a reward system from human demonstrations that can be used to train a mobile robot \cite{mirsky_prevention_2020}. Inverse reinforcement learning was then developed with Maximum Entropy Reinforcement Learning which mitigated the algorithm's reliance on the quality of the data provided because of its use of variability to allow for the algorithm to explore its own fundamental principles it picks up from the demonstrator\cite{ziebart_maximum_2008, kretzschmar_socially_2016}. Another development to the Inverse Reinforcement Learning approach is Adversarial Inverse Reinforcement Learning, an approach that simultaneously creates the reward and value function with the aim of creating a robust reward system that can be utilized in different environments \cite{fu_learning_2018}. Maximum Entropy Reinforcement learning is the preferred method because it allows the algorithm to explore on its own and adversarial inverse reinforcement learning is computationally intensive. Even with Maximum Entropy Reinforcement Learning, there is still the issue of the variability being too extreme, leading the algorithm to stray too far from demonstrator data. Deep Learning helps mitigate this variability because it allows us to create a set framework for maximum entropy to then be applied. This has been applied to human social navigation behavior and tested in simulations \cite{fahad_learning_2018}; however, this still lacks real-world testing.

\subsection{Data collection for stimuli-driven human navigation}


Some researchers will also turn to pre-existing data sets for data like the Human-Robot interaction data collection for vision-based Navigation(HuRoN) database \cite{hirose_sacson_2023} and the Multi-Modal Social Human Navigation Dataset (MuSoHu) \cite{xiao_multi-modal_2023}, both of which collect data from real-world interactions. They struggle from being hard for researchers to replicate if they want to add to the data set or modify how the data is collected.

Another approach to data collection is curriculum-based data collection has been used to help demonstrators better understand what is asked of them by the researchers as the tasks they need to accomplish become progressively harder. It has also been used in the training of algorithms mimicking the way the human brain learns and remembers new information \cite{soviany_curriculum_2022}. Curriculum-based data collection and training have been used in tandem with a deep reinforcement learning model to help reduce bias and streamline the data collecting and training process of the algorithm. This was done with human navigational data, and this work did not address inverse reinforcement learning \cite{asselmeier_evolutionary_2023}. While curriculum-based data collection is effective, researchers have to replicate this environment in the real world, which can be extremely costly and time-consuming. 

A new data collection approach is game-based data collection. This approach grew in popularity as mobile robots are able to emulate human behavior present in games and demonstrations \cite{bentivegna_humanoid_02, argall_survey_09}. Because of this, many researchers have utilized games to collect data for their research, leveraging the competition they bring and implementing what is called a game theoretic approach \cite{oguzhan_data_23, akcin_data_23}. Through a game-based data collection method, researchers are able to specifically curate the data they want to collect while avoiding the costs that replicating the environment would have.

\begin{figure}[!htbp]
     \centering
     \includegraphics[width=\columnwidth]{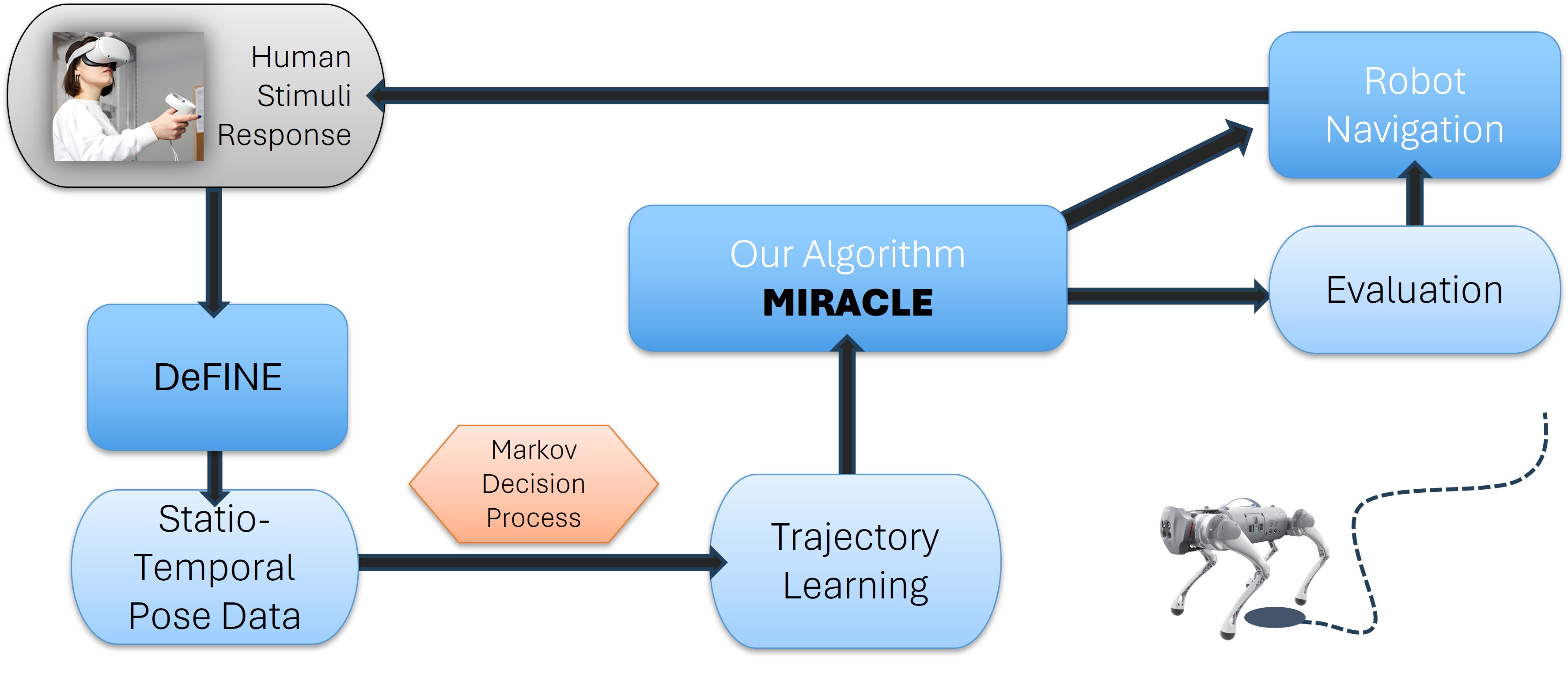}
     \caption{\textit{The figure displays the overall workflow of our MIRACLE pipeline. We utilize gamified learning to gather data regarding people’s responses to stimuli in a controlled setting. To limit the impact of demonstrators’ bias, we developed a deep maximum entropy inverse reinforcement model to benefit from the variance that maximum entropy embeds into the algorithm while using deep learning to ensure that the algorithm does not stray too far from its intended goal}.}
     \label{fig:workflow}
 \end{figure}

\section{Problem Formulation}

In this section, we introduce our newly developed socially-aware navigation algorithm for mobile robots, \method. To comprehend its intricacies, it's crucial to first understand its foundations in the realm of Markov Decision Processes (MDPs). Let's break this down.

Imagine a mobile robot working within a three-dimensional space, akin to our daily environment. This robot's movements and decisions can be represented by what we call an MDP.

\textbf{MDP Elements:}

\begin{itemize}
    \item \textbf{States \(S\):} These represent all possible positions or situations our robot might find itself in. In our 3D environment, we denote this as \(S=\{s_{1}...s_{n}\}\), where 'n' is the total number of such possible states.

    \item \textbf{Actions \(A\):} At any given position or state, our robot can take certain actions, like moving forward, turning, or stopping. This is given by \(A = \{a_{1}...a_{p}\}\), where 'p' denotes the total number of possible actions.

    \item \textbf{Transitions \(T(s_{i}, a_{j}, s_{i+1})\):} Think of this as the robot's rulebook. It tells the robot the probability of moving from one state to another, given it takes a certain action.

    \item \textbf{Discount Factor, \(\gamma\):} This factor, which is between 0 and 1, decides how much importance the robot gives to its future rewards. A higher value means the robot cares more about long-term rewards.

    \item \textbf{Reward Function \(R(s_{i})\):} Each time the robot takes an action from a state, it gets a 'reward' or 'penalty'. This function calculates that value, helping the robot understand how good or bad its decision was.

    \item \textbf{Utility Function \(U(s_{i})\):} This combines the reward for the current state and the highest possible reward for future actions. It guides the robot in choosing the best possible action to take next.
\end{itemize}

Using these elements, the machine learning model can learn the most beneficial set of actions, or 'policy', the robot should follow to reach its destination swiftly while avoiding obstacles.

\textbf{Defining the Problem:}

Given a set of policies \(D = \{Q_{1}...Q_{n}\}\), our challenge is to pinpoint the best reward function, \(R(s_{i})\). This would ensure our robot efficiently avoids obstacles and people, reaching its destination as swiftly as possible, taking into account the transition probabilities.

\section{Curriculum-based Maximum Entropy Deep Inverse Reinforcement Learning}
In this section, we discuss our proposed curriculum-based deep inverse reinforcement learning algorithm, \method. We also discuss the underlying gamified learning framework on which the proposed algorithm is based along with the changes we made to the original framework.

\subsection{Inverse Reinforcement and Curriculum Learning Model (\method)}
First, a deep learning model is used to learn the demonstrator's stimuli-based navigational behavior. Then, a maximum entropy reinforcement learning model is applied to the deep learning model's optimization function to embed variance into the policy the algorithm uses and allow the algorithm to explore for itself. 


The Deep Learning network used consists of an input layer represented by $x$ with 2 neurons representing x and y coordinates. This input layer is then connected to two hidden layers, each consisting of $128$ neurons with the activation functions for each layer given by,  
\begin{equation}
    h_1 = ReLU(W_1*x+b_1)
\end{equation} and \begin{equation}
     h_2 = ReLU(W_2*x+b_2) \,,
\end{equation} 
where $W$ is the weight matrix and $b$ is the bias vector corresponding to each hidden layer. The output of this network is then defined as:\begin{equation}
    y = W_3*h_2+b_b \,.
\end{equation}
The core Maximum Entropy Inverse Reinforcement Learning process is implemented to optimize the reward function when iterating over training epochs as shown in Algorithm~\ref{alg:IRL}. Each state-action pair is represented by $a|s$, in the trajectory. The maximum entropy loss (MEL) is calculated by:
\begin{equation}
    MEL = -\sum_{a}p(a|s)log(p(a|s)) \,.
\end{equation} 

The alignment loss (AL), i.e., the difference between the demonstrators' policy and the learned policy is then calculated as, 
\begin{equation}
    AL = -\sum_{s}f(s)\sum_{a}p(a|s)log(p(a|s)) \,,
\end{equation}

where $f(s)$ is the state visitation frequency calculated as:
\begin{equation}
    f(s) = \dfrac{1}{NT}\sum_{i=1}^{N}\sum_{t=1}^{T}s_{i,t}\,.
\end{equation}
Here, $N$ is the total number of trajectories, $T$ is the length of each trajectory, $s_{i,t}$ is the state vector at step $t$ in trajectory $i$. The maximum entropy objective (MEO) is then calculated as:
\begin{equation}
    MEO= MEL + AL \,.
\end{equation} 

Given the MEO, the gradient is then calculated using automatic differentiation, and the model parameters are updated using Legacy ADAM ~\cite{kingma2014adam} for the learning rate $\alpha$. We chose to use Legacy Adam because it adapts its learning rate and parameters based on the scale of past gradients. Legacy Adam alsos integrates with tensor flow and keras which are the libraries we used to code the algorithm.



\begin{algorithm}
\caption{\method}
\label{alg:IRL}
\begin{algorithmic}[1]
\Require
\Statex $humTraj$: Example human trajectories
\Statex $trajLen$: Length of each trajectory
\Statex $sDim$: Dimension of the state space
\Statex $lr$: Learning rate for optimization
\Statex $epochs$: Number of training epochs
\Ensure
\Statex $irlModel$: Trained Inverse Reinforcement Learning model

\Function{createIrlModel}{$inputDim, outputDim$}
    \State $model \gets$ initSeqModel() \Comment{Initialize a sequential neural network model}
    \State Add a dense hidden layer with ReLU activation and input dimension $iDim$
    \State Add another dense hidden layer with ReLU activation
    \State Add an output layer with linear activation
    \State \textbf{return} $model$
\EndFunction

\Function{calStateFreq}{$humTraj, sDim, trajLen$}
    \State $sCnts \gets$ initZeroArray($sDim$) \Comment{Array for state counts}
    \For{$traj$ \textbf{in} $humTraj$}
        \For{$(state, \_)$ \textbf{in} $traj$}
            \State $sCnts \gets$ updateStateCnts($sCnts, state$) \Comment{Update state counts}
        \EndFor
    \EndFor
    \State $sFreq \gets$ normStateCnts($sCnts, humTraj, trajLen$) \Comment{Normalize by total states visited}
    \State \textbf{return} $sFreq$
\EndFunction

\Function{maxentIrlOpt}{$model, humTraj, sDim, lr=lr, epochs=epochs$}
    \State $trajLength \gets$ lenOfTraj($humTraj[0]$) \Comment{Get length of a single trajectory}
    \State $optimizer \gets$ initAdamOpt($learningRate=lr$) \Comment{Initialize the Adam optimizer}
    \For{$epoch \gets 1, epochs$}
        \State $totalLoss \gets 0$ \Comment{Initialize total loss for epoch}
        \State $sFreq \gets$ calStateFreq($humTraj, sDim, trajLen$)
        \For{$traj$ \textbf{in} $humTraj$}
            \For{$(state, \_)$ \textbf{in} $traj$}
                \State \textbf{Using} GradientTape:
                    \State $pref \gets$ getModelPref($model, state$) \Comment{Get model's preferences state's}
                    \State $prob_h \gets$ softmax($pref$) \Comment{Preferences to a probability distribution}
                    \State $max_{el} \gets -\sum(prob_h \cdot \log(prob_human))$
            \EndFor
        \EndFor
    \EndFor
\EndFunction
\end{algorithmic}
\end{algorithm}

The pseudocode, as seen in Algorithm~\ref{alg:IRL} represents this process as follows. First, the inverse reinforcement learning model is created based on the input parameters with two hidden layers of ReLU activation and one output layer with linear activation. This is followed by the calculation of the state frequency function that creates an array of state counts and calculates the state frequency with them. Then the maximum entropy inverse reinforcement learning optimization function in which Legacy Adam and loss calculations are created, and the principle of maximum entropy is embedded into the neural network.




\subsection{Delayed Feedback-based Immersive Navigation Environment (DeFINE)}

To deploy the \textit{\method} algorithm effectively, it's essential to understand human social navigation. Envision observing people in a busy public place, keenly noting how they move, avoid collisions, and choose paths. Real-world observations can be tricky due to uncontrollable variables. Hence, we utilize a unique tool: a virtual reality (VR) platform named \textit{Delayed Feedback-based Immersive Navigation Environment}, or \textit{DeFINE} by Tiwari et al.~\cite{tiwari_define_2021}.

\textbf{Distinctive Features of DeFINE:}

\begin{enumerate}
    \item \textit{DeFINE} is crafted using Unity3D, a renowned platform for interactive simulations, ensuring realism and immersion.
    
    \item The "delayed feedback" mechanism ensures participants receive feedback only post their VR journey. It's akin to receiving a game score after completing all levels.
    
    \item The platform's modularity facilitates customization, enabling tailored environments, game-like challenges, and control over visual and auditory cues. Such adaptability aids in discerning the influence of various stimuli on navigation choices.
\end{enumerate}

\textbf{Relevance of DeFINE to our Research:}

\begin{itemize}
    \item \textbf{Simulating Real Challenges:} The delayed feedback mirrors real-life scenarios where immediate outcomes aren't always discernible. This feature resonates with complexities in inverse reinforcement learning.
    
    \item \textbf{User Comfort:} While some VR tools can induce motion sickness, \textit{DeFINE} ensures a comfortable experience, eliminating the risk of collecting skewed navigation data.
    
    \item \textbf{Data Integrity and Privacy:} While data collection is paramount, so is participant confidentiality. \textit{DeFINE} seamlessly logs essential data while maintaining participant anonymity.
\end{itemize}

Upon gathering adequate navigation data within the \textit{DeFINE} framework, the next objective is data transformation to a format digestible by the \textit{\method} algorithm. Specifically, positional data is converted to depict movement paths or trajectories. These trajectories undergo processing via Algorithm~\ref{alg:createHumanTraj}, refining them for enhanced algorithmic analysis. Consequently, a bridge is established between human navigation nuances and the \textit{\method} algorithm's computational capabilities.

 \begin{algorithm}
 \caption{Parse CSV File}
 \label{alg:parse_csv_file}
 \begin{algorithmic}[1]
 \Function{parse\_csv\_file}{file\_path}
     \State $df \gets \text{pd.read\_csv}(file\_path)$
     \State $pos\_data \gets df[['pos\_x', 'pos\_z']].\text{values.tolist}()$
     \State \Return $pos\_data$
\EndFunction
\end{algorithmic}
\end{algorithm}

\begin{algorithm}
\caption{CreateHumanTraj}
\label{alg:createHumanTraj}
\begin{algorithmic}[1]
\Function{createHumanTraj}{pos\_data, traj\_len, state\_dim}
    \State $num\_individuals \gets \text{len}(pos\_data)$
    \State $human\_trajs \gets []$
    \For{$\_ \text{in range}(num\_individuals)$}
        \State $traj \gets []$
        \State $init\_state \gets \text{np.array}(pos\_data[\_])$  \Comment{Initial state from XYZ positions}
        \State $state \gets init\_state$

        \For{$\_ \text{in range}(traj\_len)$}
            \State $action\_coeff \gets \text{np.random.uniform}(-1, 1, \text{size}=state\_dim)$  \Comment{Random action coefficients}
            \State $action \gets action\_coeff \times 0.1$  \Comment{Scale actions}
            \State $new\_state \gets state + action$
            \State $traj.\text{append}((state, action))$
            \State $state \gets new\_state$
        \EndFor

        \State $human\_traj.\text{append}(traj)$
    \EndFor

    \State \Return $human\_traj$
\EndFunction
\end{algorithmic}
\end{algorithm}

\section{Empirical Evaluation}
In this segment, our primary aim is to unravel our systematic approach for acquiring field data, evaluating the effectiveness of \method, and subsequently translating this learned navigation behavior into a real-world mobile robot setup.

\begin{itemize}
    \item Field Data Acquisition: Sec.~\ref{subsec:data-collection}
    \item Performance Evaluation of \method: Sec.~\ref{subsec:evaluation}
    \item Transference to Mobile Robot: Sec.~\ref{subsec:demo}
\end{itemize}

\subsection{Data collection} \label{subsec:data-collection}
\begin{figure}[!htbp]
    \centering
    \includegraphics[scale=0.24]{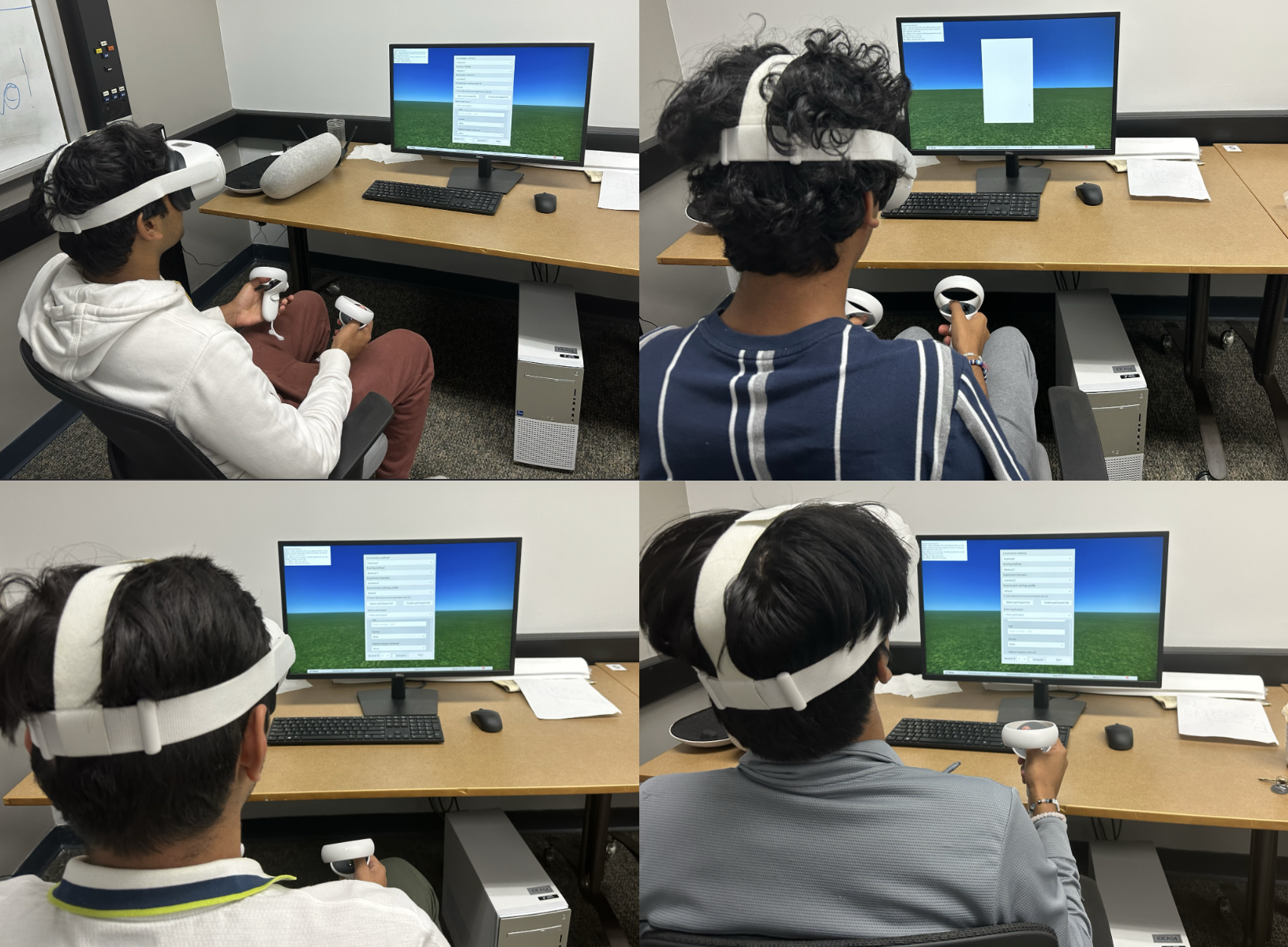}  
    \caption{\textit{Game data collection model with different individuals, each paired with a VR headset on the DeFINE game}.}
    \label{fig:VRSetUp}
\end{figure}

The heart of our empirical analysis is the DeFINE environment, shown in Fig \ref{fig:VRSetUp}. Conceptualize a simulated square room wherein an individual begins their journey from a predetermined point. Their mission? To locate and reach a concealed destination. But, there's a catch! A fly, a mere visual artifact, buzzes around the goal. This fly acts as a distraction, a noisy visual cue, challenging the participant's navigation skills.

The goal isn't to chase the fly. Instead, participants were instructed to deduce the hidden goal's location using the fly's movement. To ensure clarity, this directive was emphasized prior to the trials. The engagement doesn't end there. After each trial, the participants received a comprehensive score, considering their proximity to the target and their speed. This scoring mechanism was not merely informational; it spurred a competitive spirit. With a leaderboard in play, participants were driven to refine their strategies, evolving in their responses to the stimuli. This transformed a simple navigation task into a gamified learning experience.

For our study, 15 diverse participants were chosen, spanning different age groups, genders, and educational backgrounds. Each participant was equipped with the same briefing and navigated through the scenario 15 times. The environment was rendered on a Windows 10 system, powered by an i7 Intel CPU, ensuring consistent performance and visualization. Every movement and every decision of these participants was meticulously recorded. With 15 attempts at their disposal, their primary objective was simple - to clinch the top spot on the leaderboard.

To uphold ethical standards, all recorded data was thoroughly anonymized. Moreover, the data was timestamped, which allowed for a granular analysis of participants' decision-making processes over time.

\subsection{Performance evaluation} \label{subsec:evaluation}
Evaluating the performance of \method is crucial in understanding its efficiency and effectiveness. Here's how we went about it:

\noindent \textbf{Setup and Parameters:}
For our experiment, we chose a powerful Macbook with an M2 Max chip, ensuring robust processing capabilities and seamless execution. The algorithm was subjected to $100$ training cycles/epochs. Within these epochs, we set the learning rate, denoted by $\alpha$, to a small value of $0.001$. This controlled rate ensures the algorithm learns steadily without overshooting or missing its optimal solution.

\noindent \textbf{Trajectory Details:}
The trajectory data, a sequence capturing movement details, was specifically designed with a length of $20$ time steps. Each of these steps was incremented at a consistent rate of approximately 0.1 seconds. This granularity allowed \method to access frequent snapshots of the movement, granting it a detailed insight into the navigation strategies employed.

\begin{figure}[!htbp]
    \centering
    \includegraphics[scale=0.24]{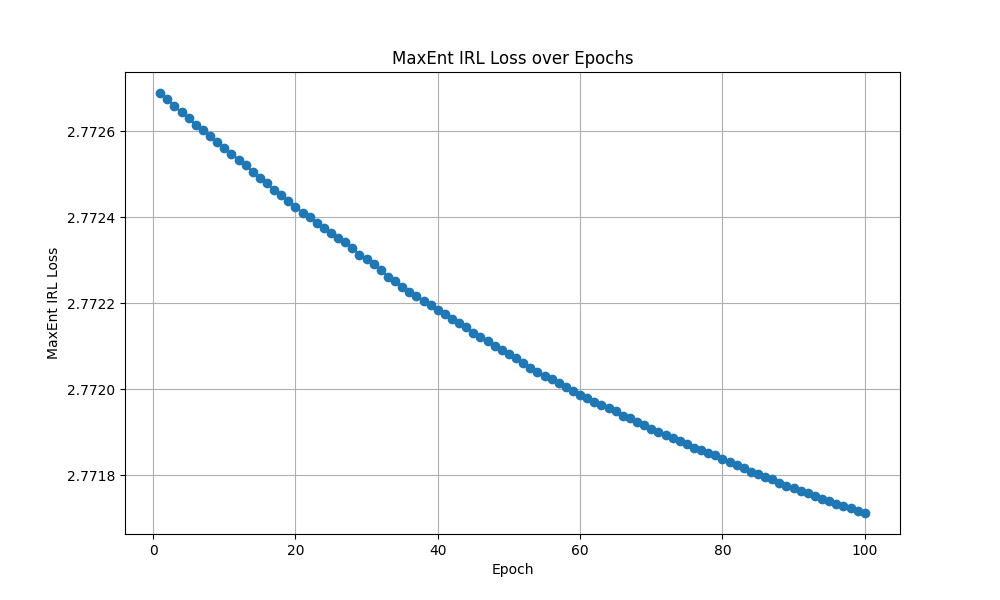}  
    \caption{The Figure displays the function optimizing its loss over the epochs of 1-100.}
    \label{fig:Results}
\end{figure}

\noindent \textbf{Results:}
Upon processing, within a reasonably large environment size of $400$ units, \method displayed an admirable convergence. The algorithm successfully reduced its error, or 'loss', to a mere $2.7717$. This loss value signifies the difference between the algorithm's prediction and the actual data. The closer this value is to zero, the better the algorithm's performance. As shown in Fig.~\ref{fig:Results}, this achievement underscores \method's capacity to effectively grasp and interpret the data it was fed.

\noindent \textbf{Implications:}
Such a low loss value, achieved within the constraints of our setup, indicates that \method can be reliably employed for interpreting navigation data. Its efficient learning and prediction capabilities make it a worthy contender for applications in diverse scenarios requiring navigation intelligence.

\subsection{Demonstration with the Quadruped Robot} \label{subsec:demo}

To truly understand how \method works in real-life situations, we decided to put it to the test using a four-legged robot called the Unitree GO1. Let's dive into this fascinating experiment.

\noindent \textbf{Choosing the Robot:}
The Unitree GO1 isn't just any robot; it's a highly mobile and agile quadruped robot, designed to maneuver through a variety of environments. Its stability and adaptability made it the perfect candidate for our demonstration.

\noindent \textbf{Setting the Stage:}
We initiated our experiment in an empty room, giving the robot plenty of space to move around. Our goal was simple: we wanted to see if the robot, powered by the \method algorithm, could locate a specific goal based solely on stimuli from its environment.

\noindent \textbf{The Stimulus:}
Instead of relying on complex signals, we used a basic yet effective stimulus: light. We positioned a light source close to our designated goal spot. This light acted as a beacon, guiding the robot to its destination. The challenge? The robot had to recognize the light as a signal to locate the goal without any other guiding instructions.

\noindent \textbf{Feedback Mechanism:}
As the Unitree robot maneuvered its way towards the light, we continuously monitored its movements. Whenever it got close to the goal, which we defined as a circle about the same size as the robot, we gave it feedback. This feedback wasn't just about whether it reached the goal or not, but also about how efficiently it did so.

\noindent \textbf{Results and Implications:}
The results were promising. The GO1 robot, using the \method algorithm, was consistently able to locate the light-source and, by extension, the goal position. This achievement was more than just a robotic feat; it was a testament to the power of deep learning combined with maximum entropy reinforcement learning.

But what does all this mean? Firstly, it shows that our DeFINE platform can effectively gather human navigation data based on stimuli, which can then be translated to robots. More importantly, it underscores the potential of robots to interpret and act on simple environmental cues, opening doors for more intuitive and intelligent robotic systems in the future.

\subsection{Discussion} \label{subsec:discussion}

DeFINE provides a controlled environment for capturing human navigation behavior in response to specific stimuli. This environment mimics real-world conditions without the inherent unpredictability and complexity. Collecting similar data in an actual physical setting would require significant logistical efforts, resources, and would introduce more variables that could skew results. DeFINE's customizable features allow for specific scenario testing. For instance, if we wish to test navigation behavior with varying light intensities or moving obstacles, DeFINE lets us modify these parameters quickly.

MIRACLE's ability to interpret and respond to stimuli, as evidenced by its navigation patterns, closely mirrors human responses under similar conditions. This similarity reinforces the algorithm's viability in applications that require human-like behavior. The observed loss value, while on the higher side, is within acceptable limits. It indicates two things: First, MIRACLE is indeed learning and adapting from the input data. Second, the deviation from the demonstrator's data suggests MIRACLE is not just overfitting or directly mimicking the data but is, instead, generating its unique responses based on its understanding.

\section{Conclusions}
In this work, we proposed the utilization of the virtual reality environment, DeFINE, to collect human social navigation data, and learn how humans react to stimuli. This data was then used to train our proposed method,\method- a combination of Deep Learning and Maximum Entropy Inverse Reinforcement Learning. The \method~algorithm was developed to be able to react to stimuli in a similar fashion as human beings such that it can be used in emergency situations, where a mobile robot equipped with the MIRACLE algorithm would be able to determine where a potential victim could be based on the observed stimuli. This method is beneficial because it utilizes a game to collect human social navigation. It also leverages the benefits of maximum entropy inverse reinforcement learning to prevent the impact of demonstrator biases on the algorithm while also leveraging the benefits of deep learning to keep the algorithm grounded.

In future works, we would like to see DeFINE used for human social navigation data. We would also like to see more research in combining maximum entropy reinforcement learning with deep learning so that algorithms can leverage the benefits of both to optimize themselves and not be skewed on demonstrator data. Beyond that, we want to see more research in the field of stimuli-driven human navigation, alongside its emergency situation applications.

\bibliographystyle{IEEEtran}
\bibliography{root}

\end{document}